%% file: main.tex
\documentclass{article}

\usepackage{microtype}
\usepackage{graphicx}
\usepackage{subfigure}
\usepackage{xcolor}
\usepackage{booktabs} %
\usepackage{todonotes}

\usepackage{hyperref}
\renewcommand{\baselinestretch}{.9991}

\usepackage[accepted]{mlsys2020}

\mlsystitlerunning{Amazon SageMaker Autopilot}

\begin{document}

\twocolumn[
\mlsystitle{Amazon SageMaker Autopilot: \\
a white box AutoML solution at scale}

\mlsyssetsymbol{}{}

\begin{mlsysauthorlist}
\mlsysauthor{Piali Das, Valerio Perrone, Nikita Ivkin, Tanya Bansal, Zohar Karnin, Huibin Shen, Iaroslav Shcherbatyi, Yotam Elor,}{}
\mlsysauthor{Wilton Wu, Aida Zolic, Thibaut Lienart, Alex Tang, Amr Ahmed, Jean Baptiste Faddoul, Rodolphe Jenatton$^*$,}{}
\mlsysauthor{Fela Winkelmolen$^*$, Philip Gautier$^*$, Leo Dirac$^*$, Andre Perunicic$^*$, Miroslav Miladinovic, Giovanni Zappella,}{}
\mlsysauthor{C\'{e}dric Archambeau, Matthias Seeger, Bhaskar Dutt, Laurence Rouesnel}{to}

\end{mlsysauthorlist}

\mlsysaffiliation{to}{Amazon Web Services}

\mlsyscorrespondingauthor{Piali Das}{pialidas@amazon.com}
\mlsyscorrespondingauthor{Valerio Perrone}{vperrone@amazon.com}

\mlsyskeywords{Machine Learning, MLSys}

\vskip 0.3in

\begin{abstract}
AutoML systems provide a black-box solution to machine learning problems by selecting the right way of processing features, choosing an algorithm and tuning the hyperparameters of the entire pipeline. Although these systems perform well on many datasets, there is still a non-negligible number of datasets for which the one-shot solution produced by each particular system would provide sub-par performance. In this paper, we present Amazon SageMaker Autopilot: a fully managed system providing an automated ML solution that can be modified when needed. Given a tabular dataset and the target column name, Autopilot identifies the problem type, analyzes the data and produces a diverse set of complete ML pipelines including feature preprocessing and ML algorithms, which are tuned to generate a leaderboard of candidate models. In the scenario where the performance is not satisfactory, a data scientist is able to view and edit the proposed ML pipelines in order to infuse their expertise and business knowledge without having to revert to a fully manual solution. This paper describes the different components of Autopilot, emphasizing the infrastructure choices that allow scalability, high quality models, editable ML pipelines, consumption of artifacts of offline meta-learning, and a convenient integration with the entire SageMaker suite allowing these trained models to be used in a production setting.
\end{abstract}
]

\printAffiliationsAndNotice{$^*$Work done while at Amazon Web Services.}  %

\input{intro.tex}

\input{related.tex}

\input{our_approach.tex}

\input{core.tex}

\input{metalearning.tex}

\input{results.tex} 
\section{Discussion}
In this work we discuss the importance of flexibility, transparency and customization when building AutoML solutions, and described SageMaker Autopilot, a large-scale industrial AutoML system.
We described the strategies adopted to create a system which provides both one-click end-to-end solutions and
the ability for expert customers to change the generated code to build their own specialized solutions. 
We evaluated SageMaker AutoPilot against several baselines on a large collection of datasets, showing the robustness 
of the system and the competitive performance obtained by the generated models. Moreover, Autopilot runs
significantly faster than other AutoML systems.

Autopilot was designed to be modular and abstract components for complexity and scalability. For instance, Autopilot’s docker architecture allows each feature engineering, algorithm, and post processing job to be run in its own docker container, yielding seamless customization. Each container is also executed on dynamically-sized hardware determined by Autopilot. These design decisions allow for multiple algorithms and frameworks, and to scale to larger datasets and problems. This has allowed us to add new algorithms, such as MLPs, and preprocess complex data types such as date and time. We plan to leverage this design to expand the diversity of algorithms, processors, and data types. Users can also bring their own preprocessors, and eventually we may even let users bring their own algorithms.

Future work also includes automating a wider range of tasks. While some tasks (such as NLP in tabular form) already work out of the box, other problems (e.g., vision, forecasting, and non-tabular NLP) can be first mapped to a tabular data format and then fed into Autopilot. Our expert-in-the-loop approach also allows users to bring in customized transformers for new type of entries, such as images or sound.
\input{acknowledgments.tex}

\bibliographystyle{mlsys2020}
\bibliography{acmart,references}
\end{document}

%% file: intro.tex
\section {Introduction}

Over the last decade, Machine Learning (ML) received an increasing amount of attention from different industries. 
Its diffusion was facilitated by the advent of cloud computing: compute resources became readily available, 
enabling scientists to explore the usage of techniques that were otherwise unaffordable for most of them. 
Today, there is an abundance of advanced algorithms and infrastructure 
that allow us to build ML-based solutions to meet business and societal needs. 
 
With the diffusion of machine learning, also the need for expert machine learning scientists and engineers grew,
in many cases creating significant problems to organizations willing to scale up their usage of these techniques.
In order to obtain good performance in real ML applications, a number of important choices should be made along the way. 
For example, how to pre-process the available data and which ML algorithm to choose.
In addition, machine learning algorithms usually have a long list of
training parameters (also known as hyperparameters) that need to be set ``just
right'' if you want to squeeze every bit of extra accuracy from your models. 
To make things worse, there are several other decisions regarding the infrastructure on which the ML applications are run which require addition choices, such as picking the compute resources to ensure 
the model can be trained while still keeping cost under control. 
 
The lack of data scientists %
and an over-abundance of data science problems are not the only problems faced in the industry. 
Even if experts are available, many of these choices are specific to the considered application and there is often the need of a 
significant number of trial and error experiments to find the optimal solution.
With a vision to reduce these repetitive development costs, the concept of automated machine learning (AutoML) has emerged
in recent years and has become a hot area of research. 
In what follows we describe the solution provided by Amazon SageMaker, designed on one hand to enjoy the advances provided by academic research, but on the other hand to take into account the realistic needs of the industry user, which is not fully catered to by state-of-the-art research papers.

\section{The full ML pipeline}
 
We start by describing the different steps involved in building an ML-based application, including both the analysis and exploration steps as well as the execution of the ML pipeline.

\begin{enumerate}
    \item \textbf{Meta-data extraction:} Collect statistics related to the entire dataset and related to individual columns. Example features for individual columns include: fraction of missing entries, percentiles, skewness, correlation with the target. Example features for the entire dataset are the number of rows/columns, the distribution of columns among feature types, landmark features~\cite{vanschoren2018meta} that identify performance of some ML model on a data subsample.  
	This stage is useful for gathering insights about the data, discovering faulty inputs and guiding the AutoML engine towards better solutions. \label{itm:metadata}
    
    \item \textbf{Dataset analysis:} Given the extracted meta-data, provide an analysis to guide the system: detect the problem type (e.g., regression or classification); detect the column schema (e.g., numeric, categorical, natural language, datetime). Based on a combination of good practices and a meta-learned model,
	provide guidance to the optimization problem in the following stages by limiting the search space and providing priors. \label{itm:analysis}
    
    \item \textbf{Algorithm training and evaluation}: This component applies a specific machine learning pipeline; its input is the entire recipe
    of the type of feature processing, the algorithm and its hyperparameters. \label{itm:train}

    \item \textbf{Pipeline tuning:} Based on the guidance provided by \ref{itm:analysis}, invoke multiple ML pipelines via the component 
    \ref{itm:train}, in order to achieve a diverse collection of well performing pipelines and corresponding models. \label{itm:hpo}
\end{enumerate}

A summary of these steps is depicted in Figure~\ref{fig:system-architecture}. At the core of our system is a meta-model that given meta-data extracted from the input dataset (steps \ref{itm:metadata}), outputs an ML pipeline, meaning the full recipe for how to solve the ML problem. This meta-model, described in Step~\ref{itm:analysis}, is intended to mimic the work that is typically done by a scientist. In an analogy to an ML problem, the meta-model is trained with a train collection of datasets and evaluated on a separate test collection of datasets. The analogy unfortunately stops here. The two difficulties lie in the fact that (1) the size of the collection is limited - we cannot hope to have millions of examples, and (2) the fundamental assumption of ML is that the distribution of the train and test collection is identical. Finding a collection of datasets that truly represents the inputs a system would encounter in the wild is very challenging as public datasets tend to be too clean and processed. To mitigate this issue we restrict the meta-learning, incorporate manual guidance, and invest in highly informative meta-features that allow robustness to new inputs and a small collection of training data.

In addition to guiding the training of the meta-model of Step~\ref{itm:analysis} we reduce the complexity of its task. Rather than requiring an output for a single pipeline, we allow it to provide a limited search space over pipelines. In steps \ref{itm:train} and \ref{itm:hpo} we search over this limited search space using techniques developed for the {\em Combined Algorithm Selection and Hyperparameter optimization (CASH)} problem.

The explicit decomposition, both conceptually and in the system design, between the meta-model providing a search space and the following steps of finding a pipeline inside that search space is exactly what is needed for a white-box solution. In a non-negligible fraction of cases, the 1-click solution does not work. This can happen because some feature was not processed correctly, data was too skewed, or some business logic is required to guide the ML process. When this happens, a data scientist can start the work from the output of the meta-model. This output contains the analysis of the data and guidelines for steps  \ref{itm:train}, \ref{itm:hpo} in a python notebook that can be edited, and thereby further guided. This means that even in this scenario where the 1-click solution did not work, the data scientist saves valuable time by not having to worry about the details and fixing only the parts the require attention.

We finally note that prediction quality is not the only target metric to be maximized. Real-life scenarios impose several constraints, such as limited memory usage, low prediction latency, and
cost of each prediction. Moreover, since some of the operations required to use
the model in a real-world application can be complex, and model deployment can eventually be repeated 
over time or on a number of different machines, it is important to also automate the deployment procedure. This prevents mistakes and mitigates the development cost. For this reason, we expose not a single ML pipeline but multiple ones, allowing a diverse set of models w.r.t.\ the metrics defined.

%% file: related.tex
\section{Related Work}

A wide range of AutoML systems for tabular data are available today, both as open source academic packages and industrial cloud-based solutions. Most of these adhere to the CASH paradigm, as does SageMaker Autopilot: a vast search space of models and model-dependent hyperparameters is searched for a single best model, or an ensemble of diverse top-performers. CASH is based on well-studied model selection and hyperparameter optimization methodology, yet can be time-consuming in practice. A more detailed overview of automatic machine learning is provided in the surveys of \citet{zoller2019survey} and \citet{he2019automl}.

{\bf Academic frameworks}. AutoWEKA~\cite{thornton2013auto} runs Bayesian optimization on top of WEKA models and supports ensembling. Auto-sklearn~\cite{feurer2019auto} has dominated a number of AutoML competitions. It selects base models from the scikit-learn ML library~\cite{scikit-learn}, incorporates multi-fidelity HPO~\cite{falkner2018bohb} as well as ensembling~\cite{caruana2004icml}. Among its innovations are meta-learning in order to warm-start hyperparameter search and careful running time management. H2O AutoML~\cite{pandey2019h2o} employs stacking along with bootstrap ensembling, with hyperparameters tuned by random search. It is frequently used in Kaggle competitions. TPOT~\cite{olson2019tpot} extends AutoML over data preprocessing and feature pipelines, employing genetic algorithms to search randomly assembled candidate pipelines and employs stacking. While more general than many frameworks, TPOT can be expensive to run. AutoGluon-Tabular~\cite{agtabular} uses multi-layer stacking with $K$-fold bagging over gradient boosting, residual MLP, linear and KNN base learners, whose hyperparameters are set to defaults. It often outperforms other frameworks given the same time budget, but final models have a large prediction latency. It is part of the AutoGluon system~\cite{auto_gluon}, which supports distributed asynchronous hyperparameter optimization for a range of other ML problems (image classification, object detection, natural language processing). Notable other AutoML frameworks are TransmogrifAI~\cite{transmogrifai_automl}, Auto-Keras~\cite{jin2019autokeras}, GAMA~\cite{gijsbers2019gama} and Hyperopt-sklearn~\cite{bergstra2015hyperopt}.

{\bf Industrial cloud-based solutions} These include Google AutoML Tables~\cite{google_automl}, H2O Driverless AI~\cite{h2o_automl}, DataRobot~\cite{datarobot_automl}, Microsoft Azure ML~\cite{microsoft_automl}, Darwin AutoML~\cite{darwin_automl}.

%% file: our_approach.tex
\section{Our approach}\label{sec:our-approach}
The philosophy behind SageMaker Autopilot is to offer a white-box AutoML solution, with the vision to democratize machine learning. It aims to make the power of ML broadly accessible, without requiring users to become experts in this field first. Autopilot is guided by several design principles:
\begin{enumerate}
    \item \textbf{Educative, transparent and repeatable}\\
        Autopilot demystifies ML for the end-user with little to no expertise in the
        field and  serves as a good starting point to apply the ML prediction power 
        to business problems. Autopilot helps to understand the real
        value of ML in specific scenarios avoiding the more expensive and thus risky
        alternative of hiring a professional data scientist. 
        In addition, its transparency serves an educational role: the generated code for the ML pipelines lets users educate
        themselves step-by-step, transformer-by-transformer, 
        and be guided through all the stages of the ML development process. 
        Ease of reproducibility and logging let users play with the 
        system and see how the input data characteristics, such as
        adding more columns or records to the table, can
        change the quality of the generated ML model.
    \item \textbf{Ability to allow an expert in loop}\\
        Autopilot positions itself as a one-button white box solution. Users can
        choose to trust the system entirely and rely on automatic decisions by only
        observing the intermediate artifacts, such as generated pipelines. 
        However, if necessary, expert can be looped into the process, by
        overwriting some suggestions of the system as well as directly editing
        the code of the generated pipelines. For example, users can decide to
        introduce their own embedding algorithms as part of the pipeline. 
        White-boxing all the generated code lets customer keep improving the
        pipeline on-site or in a more isolated environment. The benefits of this approach are also discussed in \citet{Wang2019}.
    \item \textbf{Meant for users with different expertise}\\
        Autopilot benefits ML savvy users by taking over the heavy lifting
        tasks of cleaning and preparing the dataset as well as engineering the features.
        Trained pipelines generated by the Autopilot are often used by experts as a baseline
        solution, which is further improved manually. Non-experts benefit from this
        guided process with most decisions made automatically.
    \item \textbf{Production-level reliability and scalability} \\
        While these are essential properties for any ML service nowadays, they
        are especially crucial for the non-expert user, which has little
        understanding of what can potentially go wrong and which knobs might need
        adjustments. 
    \item \textbf{Regression and classification problem type}\\ 
        Autopilot covers the two most common business-driven machine learning
        problems on tabular data: regression and classification.  
        To accommodate for non-expert users, Autopilot can detect the problem type
        automatically and adjust quality metrics accordingly. 
\end{enumerate}

%% file: core.tex
\section{Core functionality}
\begin{figure*}[]
    \centering
    \includegraphics[width=0.9\textwidth]{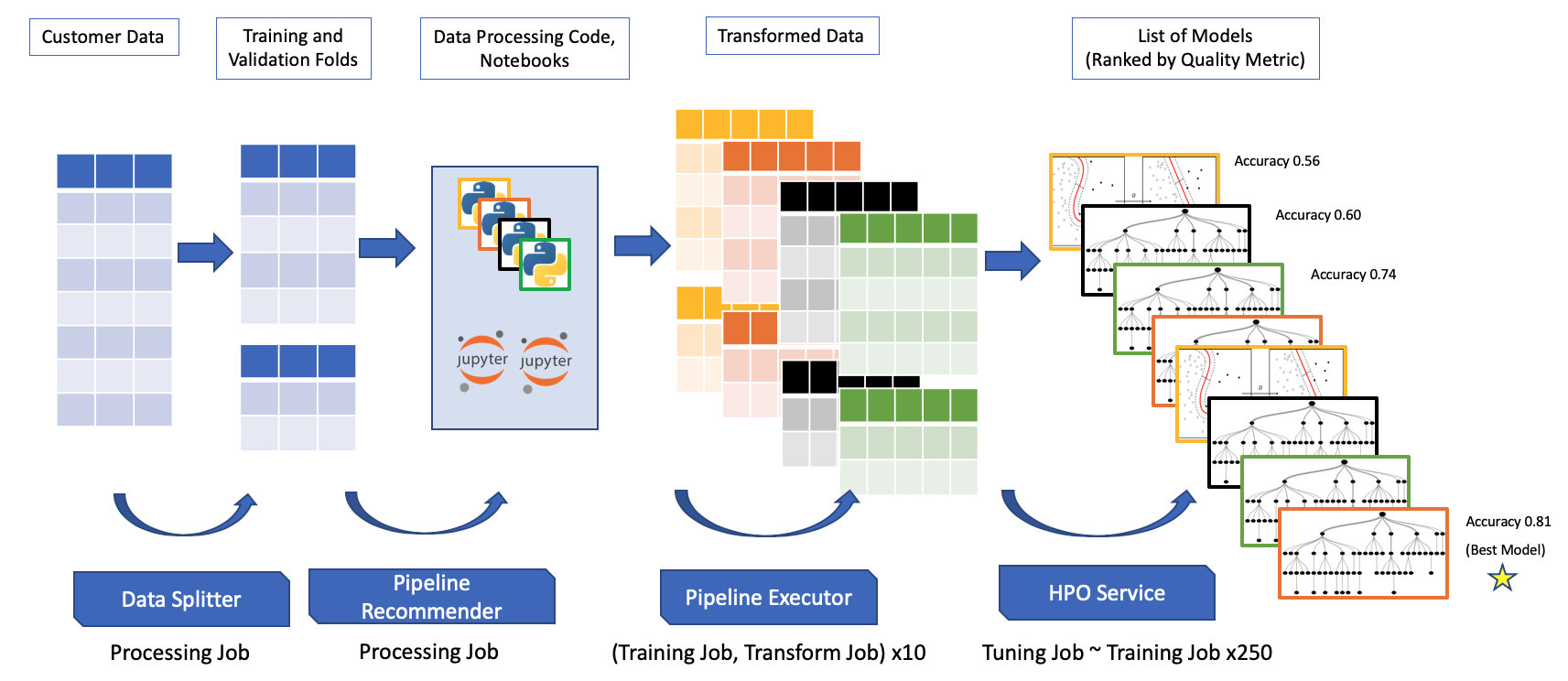}
    \caption{A high-level view of the job components that Autopilot executes while processing a customer request.} 
    \label{fig:system-architecture}
\end{figure*}

Amazon SageMaker Autopilot allows customers to quickly build classification and regression models without expert-level machine learning knowledge.
To use Autopilot, customers issue a request that includes the following information:

\begin{itemize}
	\item S3 path\footnote{Amazon Simple Storage Service (Amazon S3) is an object storage 
		service that offers industry-leading scalability, data availability, security, and performance.} 
	to a CSV file; %
	\item the name of the target column to predict;
	\item S3 location where output artifacts should be placed.
\end{itemize}

\noindent Customers can optionally specify other parameters of the job, such as
the problem type and the computational budget.\footnote{If the problem type is not specified, Autopilot
detects it automatically.}
 
Autopilot produces up to 250 consumable and ready-to-deploy models representing 
the entire ML pipelines. They can be sorted by various attributes such as the prediction accuracy. 
Tangibly, Autopilot produces two kinds of artifacts:
(1) artifacts to let customer inspect, modify and interactively re-generate the candidate models: 
\begin{itemize}
	\item A data preprocessor Python module using Scikit-Learn;
	\item A SageMaker TransformJob model used to apply the associated data preprocessor module;
	\item A trained and deployable SageMaker algorithm model paired with the data preprocessor;
	\item Train and validation folds of preprocessed dataset.
\end{itemize}
(2) All intermediate and final artifacts that includes the following:
\begin{itemize}
	\item Train and validation folds of unprocessed dataset;
	\item A Data Exploration Jupyter Notebook highlighting statistics about columns in the original dataset;
	\item A Candidate Definition Jupyter Notebook that describes each
            candidate model, and allows the customer to easily customize and
            re-deploy any or all of them.
\end{itemize}

\subsection {Internal workflow}
Autopilot consists of two primary phases: candidate generation and candidate exploration. 
In the candidate generation phase, Autopilot does the following: 
\begin{enumerate}
    \item[(1)] splits the input data into training and validation;
    \item[(2)] infers the problem type by analyzing the values in the target
        column (if not provided by user); 
    \item[(3)] generates a custom python module containing
code to transform the input data in a up to 10 possible ways; 
    \item[(4)] generates 10 tunable ML pipeline definitions in form of a Jupyter notebook, 
    referred to as Autopilot candidate definition notebook;
    \item[(5)] generates a notebook with insights about the data. 
\end{enumerate}
In the candidate exploration phase, the generated pipelines are executed with 
sophisticated hyperparameter optimization techniques to find the set of ML pipelines 
that yields optimal prediction accuracy. 
At the end, Autopilot generates inference pipeline definitions for each of the candidates 
that can be deployed  to production. 
Autopilot allows users to choose to execute only the candidate generation phase. The next section describes the underlying architecture of Autopilot 
and explains how each of the phases described above is accomplished.

\subsection{System Architecture}

Autopilot is built as an AWS  managed service that manages the stateful entity  named 
`AutoMLJob' in SageMaker's eco-system. 
It provides AWS API to manage the life cycle of the AutoMLJob. A separate ``control
plane'' component handles user-driven control actions, such as creating,
describing, and stopping the AutoML jobs. The operation of the system is handled by the "orchestrator" that executes several SageMaker compatible containers dedicated to perform specific actions in the Autopilot workflow. Figure \ref{fig:system-architecture} shows the internal flow and the system overview of Autopilot. Here, we present brief descriptions of the components of Autopilot.

\subsubsection{\textbf{The control plane}}
The control plane is a high-availability, low-latency service that processes 
API requests from the AWS client. It exposes four main APIs that can be used 
by the customers with their programming language of choice. 
It lets customers launch the job, query the status and stop the job via 
CreateAutoMLJob, DescribeAutoMLJob and StopAutoMLJob, respectively.
It provides validation of the request configuration and input data and performs authentication and authorization checks, 
based on the AWS Identity and Access Management (IAM) mechanism. 
Autopilot generates an abstract entity called "candidate", which represents an explored ML pipeline model. 
Customer can access the descriptions of all candidates via
ListCandidatesForAutoMLJob API call.

\subsubsection{\textbf{The Orchestrator and first party containers}}
 
The orchestrator component is built using AWS Step Functions, which is modeled as a sequence 
of state machines. It orchestrates SageMaker Processing Job, Training Job, 
Transform Job and Hyperparameter Tuning Job to accomplish this workflow. 

The first phase is the candidate generation. It is accomplished by running two consecutive SageMaker 
Processing Jobs, namely Data Splitter and Candidate Generator. Both are implemented as custom SageMaker compatible Docker Images, available in the AWS Elastic Container registry. 
Data Splitter checks the data sanity, performs stratified shuffling
and splits the data into training and validation.
The Candidate Generator first streams through the data to compute useful statistics.
Then, it uses these statistics to identify the problem type, and possible types of
every input column, such as numeric, categorical, and natural language.

Next, based on the characteristics of the data, an offline-trained `recommendation' model predicts at most 10 different pipelines strategies. 
Each pipeline strategy is a pair of data-processing and algorithm  
along with their hyperparameters, which can be tuned to generate a list of candidate models. 
The data-processing strategy gets translated to python scripts customized for the given dataset. 
Each such python script implements a Scikit-Learn Pipeline composed of native
Scikit-Learn transformers~\cite{scikit-learn,sklearn_api} and custom built
transformers available in an open-source python library, namely sagemaker-scikit-learn-extension \cite{ssle}. The pipeline
strategy is then translated as an ordered list of SageMaker job definitions
that are rendered as code in a Jupyter notebook, referred to as Candidate
Generation Notebook. This notebook can be executed to generate the candidate
models. The SageMaker job definitions are complete and include the resources
configuration. This phase also generates the Data Analysis Notebook that
provides insightful statistics about the dataset for the customer. This
concludes the candidate generation phase. If the customer elects to generate the
pipelines only, the state-machine stops here and marks the AutoMLJob as completed. 

The second phase is candidate exploration. 
Note that the feature preprocessing part of each pipeline has all hyperparameters
fixed, i.e., it does not require tuning. As a result, the feature preprocessing step can be done prior to running the hyperparameter optimization job. To accomplish it, 
each python script code for data-processing is executed inside a SageMaker framework 
container as a training job, followed by transform job. 
It outputs up to 10 variants of transformed data, so that the algorithms for each
pipeline are set to use the respective transformed data. 
All algorithms are optimized using SageMaker Hyperparameter Optimization. 
Up to 250 training jobs are selectively executed to find the best candidate model. 

\subsubsection{\textbf{Generated Artifacts and customization}}
As mentioned in the previous section, one of the key advantages of Autopilot is transparency. It is crucial to provide details of the generated artifacts, particularly the python scripts for data-processing, the
Candidates Definition Notebook, and how these can be inspected and customized to inject expert knowledge. As shown in Figure~\ref{fig:system-architecture}, all the
generated artifacts are available in the customer's S3 bucket. 

\begin{figure}[]
	\centering
	\includegraphics[width=0.45\textwidth]{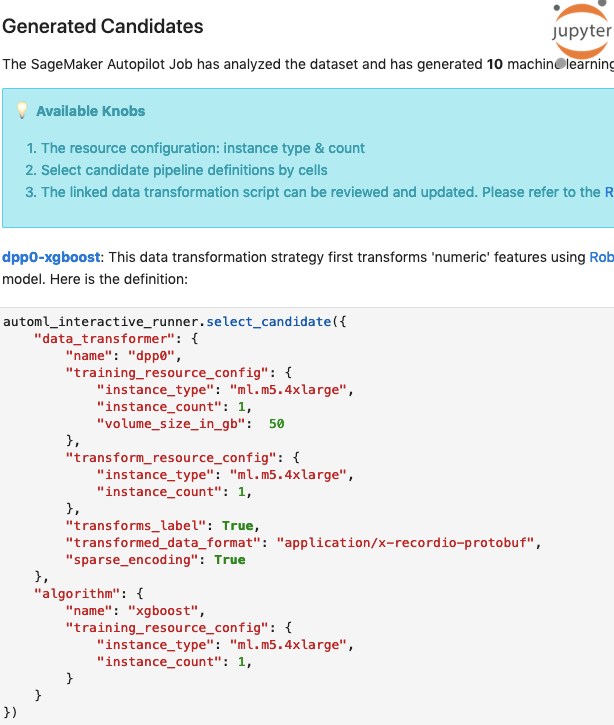}
	\caption{Snippet of Candidate Definition Notebook generated by Autopilot that describes the ML pipelines that can be modified and re-run. }
\label{fig:nb}
\end{figure}

The Candidate Definition Notebook consists of the SageMaker job definitions that are executed. The notebook can be fetched
from AWS S3. As a first step, it downloads the python scripts from S3. This allows customers to inspect and customize each of the  data-processing scripts
locally. Figure~\ref{fig:nb} shows the definition of one of the pipelines. It can be seen that it provides an intuitive description of each of the used pipelines while exposing their parameters and configuration. This allows customers to select the pipelines to explore, modify the feature and label processing techniques used in the pipeline, and specify the resource configuration for each stage of the pipeline. This also makes it possible to customize the algorithms' configurations, including static hyperparameter values, choice of the tunable hyperparameters and their respective ranges, as well as the metric to optimize for.

So far we have described how Autopilot provides a solution for customers with varying
degrees of expertise: a reliable 1-click solution  for those who are
happy with an out-of-the-box tool, and all the knobs to tweak the pipelines for experts. This is a key differentiator and adoption driver, as confirmed by feedback from a number of users. An example is Domo, who wrote in their press release: ``Domo leverages Amazon SageMaker Autopilot to make it easy to automatically train and tune ML models needed to predict outcomes while giving customers full control and visibility into their models.''\footnote{\url{https://www.domo.com/news/press/domo-announces-support-for-amazon-sagemaker-autopilot}}.

\begin{figure}[]
	\centering
	\includegraphics[width=0.45\textwidth]{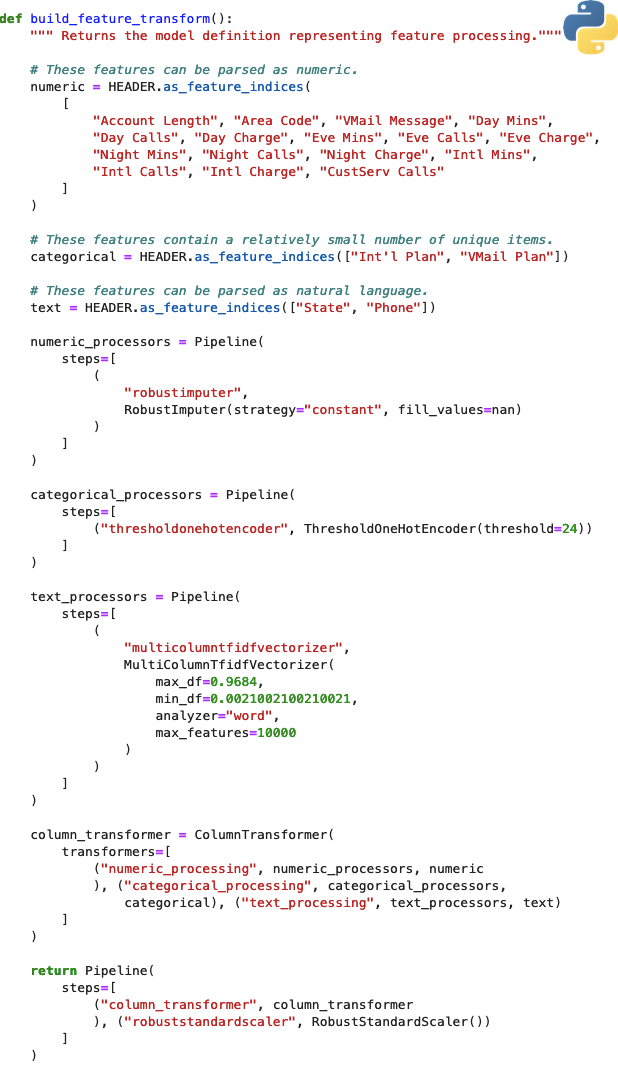}
	\caption{One of the data-processing python scripts generated by Autopilot for the given dataset. It shows the SKLearn Pipeline representing the transformation for each feature in the input dataset.}
	\label{fig:code}
\end{figure}

%% file: metalearning.tex
\section{Pipeline selection process}\label{sec:metalearning}

\begin{figure}[]
    \includegraphics[width=0.45\textwidth]{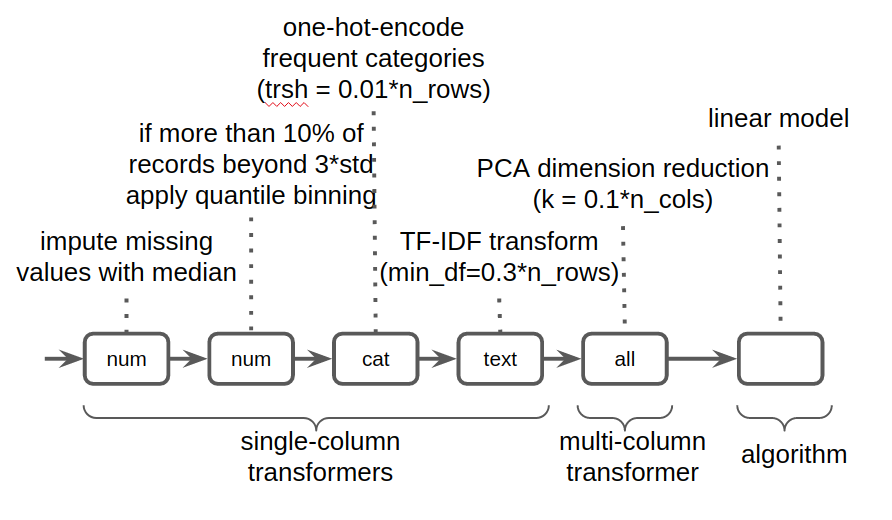}
    \caption{Example strategy including three single column transformers, one
    multi-column transformer and the algorithm. Each single-column transformer
preprocesses all columns of the specified type (numeric, categorical of text).}
    \label{fig:strategy_example}
\end{figure}

In Autopilot we differentiate between strategies and pipelines. A strategy is more tied to  the content and characteristics 
of the dataset, includes if-else statements and symbolic hyperparameters~\cite{van2018meta}. For example, consider two strategy transformers: 
\begin{enumerate}
    \item[(a)] If an input column is numeric and there are more than $X_1$ entries beyond three
        standard deviations $\rightarrow$ perform a quantile transformation with $X_2$
        bins;
\item[(b)] If the number of columns is larger than $X_3$ $\rightarrow$ apply PCA for dimensionality reduction
    with $k = X_4 \cdot $ (number of columns);
\end{enumerate}
Here, $X_{1,2,3,4}$ are the hyperparameters of the strategy. On the other hand, pipeline transformers can be seen as a realization of the strategy transformer: 
\begin{enumerate}
    \item[(a)] Apply quantile transform with $X_2$ bins to the $13$-th column;
    \item[(b)] Apply PCA dimension reduction with $k = 4$.
\end{enumerate}
The high-level structure of the strategy can be seen as a sequence of single-column transformers followed by a sequence of multi-column transformers.  An example strategy is depicted in Figure~\ref{fig:strategy_example}.

The choice of transformers from the pool of predesigned ones 
and its internal settings can be seen as
hyperparameters (HP) of the strategy: quantile binning or log transform
for numeric columns, \mbox{XGBoost} or Linear Learner, number of components in
PCA, maximum tree depth in XGBoost, and many others. Hyperparameter optimization (HPO) is traditionally considered as the most expensive
part of the ML development cycle, as each HP configuration trial includes training
of the algorithm on the entire dataset from scratch. Most widely adopted HPO
engines (SMAC~\cite{hutter2010sequential}, ParamILS~\cite{hutter2009paramils}, BOHB~\cite{falkner2018bohb}) are based on Bayesian
Optimization (BO), where a probabilistic surrogate model that maps HP configurations
to a loss is used to draw the next HP configuration to try. Initializing the surrogate model takes a portion of the optimization budget and is performed via Random Search HPO. 
BO is superior to Random Search HPO when the HP search space is small, as it was shown in practice by optimizing HP for the algorithms XGBoost or MLP.   

In our application, feature preprocessors introduce a large number of HPs in
addition to those from the algorithm, and many of these HPs are conditional. In such settings, BO fails to initialize the surrogate model and performs on par
with Random HPO. Both methods require high budgets to achieve reasonable
performance. Applying either of them on-the-fly for a new customer's dataset is
prohibitively expensive. We address this problem by splitting the set of HPs into 
two: feature preprocessing and algorithm HPs. We pre-optimize the feature preprocessing HPs in an offline manner 
and optimize the algorithm HPs online using BO. 
This approach lets us leverage the power of transfer learning and thus keep the HPO cost low. 

\subsection{Offline optimization}\label{sec:offlineopt}

It was previously shown that default hyperparameters often outperform random HPO when the number of HPO iterations is low. We adopt the zero-shot learning approach first 
introduced in \citet{wistuba2015learning, wistuba2015sequential} and aim to find a set of $k=10$ strategies such that, for a given customer's dataset, at least one of them performs well.  

To this end, we perform an offline optimization as follows: 
\begin{itemize}
    \item[-] First, we build a performance table $P$ by evaluating $B$ randomly chosen
        HP configurations on $D$ datasets, such that $P_{ij}$ is the loss of $i$-th
        HP configuration on $j$-th dataset.
    \item[-] Second, we find $k = 10$ HP configurations that minimize loss on all datasets as a group:
        \vspace{-0.1cm}        
        $$\min_I \left(\sum_{j\in D} \min_{i \in I}P_{ij}\right),$$
        where $I$ is a subset of $k$ strategies from $B$ randomly chosen
        earlier.
\end{itemize}

\subsection{Online optimization}

After recommending the feature preprocessing HPs in an offline manner, we form a
set of ML pipelines to tune. In this way, the tunable HPs are only related to the
ML algorithms. Given a fixed number of HP evaluations as budget, the natural question that arises
is: how to allocate the tuning budget across pipelines in a way that maximizes the final validation performance of the best pipeline?
One could equally distribute the budget to every pipeline. However, this may lead to
wasting too many resources on non-promising pipelines. Thus, we formulate the
budget allocation as a bandit problem where the reward of each pipeline is the
best observed validation performance so far. To this end, we found out that the simple
\mbox{$\epsilon$-greedy} algorithm \cite{sutton1998introduction} - 
where with $\epsilon$ probability, we explore a random pipeline, and with $1 - \epsilon$ probability 
we exploit the pipeline with the best reward - works better than other more complicated bandit
algorithms, such as EXP3 \cite{auer2002nonstochastic} and Rotting bandit
\cite{seznec2020single}.

To make our bandit-based pipeline selection more robust, before running \mbox{$\epsilon$-greedy} we start with an exploration phase where we 
allocate budget for the HP evaluations randomly. Our system only moves to \mbox{$\epsilon$-greedy} when the following two conditions are met:

\begin{itemize}
\item The system has \textit{suggested} 5 HPs for every pipeline. This is to ensure that even if some pipelines are slow, 
we will still observe their performance on the first 5 HP evaluations when they are finished.
\item One pipeline has \textit{finished} 5 HP evaluations. This is to ensure that when some pipelines are very slow or keep failing, 
we could still move to \mbox{$\epsilon$-greedy} phase.
\end{itemize}

Note that we could simply require every pipeline to finish 5 HP evaluations before 
moving to \mbox{$\epsilon$-greedy}. But this would make our system slow, and would pose the risk of not moving 
to \mbox{$\epsilon$-greedy} at all. After the budget has been allocated for every pipeline, we will use either random search or BO to suggest HP candidates independently 
for each pipeline. Currently, we require 5 finished HP
evaluations before using BO so that our model can be trained properly in BO.
 
The success of our pipeline selection algorithm relies on how revealing the first 5 random HP evaluations are with respect to that pipeline's performance. 
Surprisingly, with only 5 random HP evaluations before moving to \mbox{$\epsilon$-greedy}, our algorithm can successfully identify 
the best pipeline on more than 50\% of datasets that we have  evaluated; for around 80\% of our datasets, our algorithm's choice is among the top 3 pipelines. 
The success rate will be further boosted by leveraging the zero-shot approach as described in Section~\ref{sec:offlineopt}.
Instead of using 5 random HPs for an ML algorithm, we will use the learned 5 zero-shot HP configurations. 
This will significantly increase the probability that we commit to the best pipeline most of the times during the \mbox{$\epsilon$-greedy} phase. 
We refer the reader to \citet{winkelmolen2020practical} for more details on the way we find zero-shot HP configurations for an ML algorithm. Knowledge of optimal HPs could also be transferred across different datasets in future work~\cite{perrone_multiple_2017,Perrone2018,Perrone2019,Salinas2020}. 
Several ML applications also involve optimizing multiple metrics simultaneously, such as inference latency, training time or memory consumption~\citep{Perrone2019mes, Perrone2020, Lee2020, guinet2020}, which could extend the current HPO procedure.

\section{Automation}
Beyond the core functionality presented in the previous sections, Autopilot comes with a set of additional automated choices to tailor to the user's specific dataset and task. In this section, we highlight a number of ways Autopilot adapts to the problem at hand to deliver a fully automated experience.

\subsection{Schema detection}

A schema is a mapping between columns and data types. The main use of types for machine learning is the operations and semantics of the data. In Autopilot, operations refer to higher-level feature engineering primitives (e.g., one-hot encode, impute, TFIDF). Schema and type information is used to decide which set of feature pre-processors to apply. 

As a part of the pipeline selection process, Autopilot automatically determines a column's data type based on metadata collected for that column, such as the number of numeric values, string length and number of missing values. An automatic schema detection is executed in a streaming, distributed, and parallel manner across all input columns. Because of the ambiguity of input data, a single column could be classified as two or more different types. The flexibility of the schema type detection allows Autopilot to apply several different feature transformations to the same column.

\subsection{Imbalanced data}

Imbalanced classification tasks are ubiquitous. For example, consider the problem of credit card fraud detection where the majority of transactions are legitimate
and only few are fraudulent. This skewness is challenging for machine learning systems, since these tend to focus on the majority and greatly misclassify the minority.
Autopilot automatically detects imbalanced binary classification datasets and augments the ML pipeline accordingly, resulting in significant improvements of prediction quality.

\subsection{Hardware detection}
Even for a single SageMaker training job, deciding which hardware configuration (i.e., instance type, count, and volume size) to use is not trivial, especially for non-expert users. This decision depends on many factors; some are algorithm and input dataset specific, while others are finding the right trade-off between cost, speed, and optimal utilization of allocated resources. To complicate this further, the customer is not aware in advance of the feature preprocessing steps which will be taken and that can change the original dataset characteristics, such as the number of columns, density, and even storage format.
To take the burden of this choice off the customer's shoulders, Autopilot offers an automatic hardware recommendation feature. 
This allocates appropriate computational resources for each algorithm, dataset and feature preprocessing pipeline combination with the goal of preventing out of memory errors (OOMs) while avoiding unnecessary costs associated with using larger instance types or higher number of instances. 

Some algorithms, such as Linear Learner, process input data in mini-batches and, as such, rarely fail with OOM errors. Therefore, we can use a medium-sized instance type regardless of the dataset characteristics, which offers a balance between cost and efficiency. On the other hand, XGBoost is an example of a memory-intensive algorithm that requires the entire dataset to be loaded in memory for training to take place. In this case, choosing the right instance type and count becomes more critical to avoid OOMs, while keeping a low total cost and, in case of multiple instances, a low communication overhead. Note that XGBoost's hyperparameter settings also determine the memory required for training. For example, increasing the maximum tree depth hyperparameter implies higher memory requirements. For this algorithm, Autopilot's hardware recommendation logic includes a simple linear model derived from offline benchmarks that predicts the required memory based on the dataset characteristics, such as size, density, and storage format, as well as problem type, and potential hyperparameter values. Given this prediction, it is possible to select a hardware configuration in such a way that the total memory capacity is sufficiently large to handle any, even the most memory-intensive, hyperparameter choices that are potentially explored during the HPO step.

\subsection{Scalability}

With larger dataset sizes that a user might provide, stability, runtime and cost become more important factors. One of the benefits of a managed cloud solution is readily available compute resources that scale with the needs of the customer. 
In order to make Autopilot scale, a number of aspects are considered. In the following we describe a few aspects
which greatly help increase AutoPilot's scalability.

Some of the operations that are performed in a machine learning pipeline can be parallelized. 
For example, feature extraction with each of the preprocessing pipelines 
can be done in parallel. 
Hence, for all similar operations execution is done in parallel to maximally reduce the runtime, 
as long as this does not incur extra charges to the customer. 
In a similar spirit, we found no substantial degradation in the quality of the final model when evaluating ten hyperparameter configurations
in parallel instead of a single one, as often done for sequential Bayesian optimization.

Another important aspect is the search range for the hyperparameters tuned in Autopilot. Each algorithm has a different search space, which were selected so as to allow scaling without failures. For example, the total number of nodes in a decision tree is bound by the 
number of records in the dataset, as splitting of decision tree node cannot be done if the dataset sample that corresponds to the tree node contains a single record. Other problems have been observed with datasets of larger sizes where, for example, the XGBoost models would increase in size considerably, and lead to OOMs and overfitting. This is especially the case when the maximum tree depth hyperparameter is set to some high value. Defining suitable boundaries for the search space is important not only for the quality of the model but also
to guarantee a high reliability of the system.

%% file: results.tex
\section {Results}

We evaluate the performance of Autopilot on a collection of 176 tabular datasets,
obtained mainly from public sources, such as
UCI~\cite{Dua:2019}, Kaggle~\cite{kagglewebsite}, and OpenML~\cite{OpenML2013}.  
The collection includes a wide spectrum of datasets diverse in size, content, problem type, etc.
See some descriptive statistics of the datasets chosen in Table~\ref{tab:dataset-stats}.

\begin{table}[t]
\begin{tabular}{l|r}
\textbf{Parameter}                      & \textbf{Value} \\
\hline
Regression datasets                     & 50             \\
Binary classification datasets         	& 58             \\
Multiclass classification datasets      & 67             \\
Number of rows (range)                  & $[10^3, 5\cdot10^6]$\\
Number of columns (range)               & $[2,7201]$\\
Maximum number of classes               & 2428         \\
Number of datasets with text features 	& 34             \\
Size (range in Mb)                      & [0.02,  5000]     
\end{tabular}
\vspace{0.3cm}
\caption{
    \small Parameters of the datasets collection used for benchmarks.    
}
\label{tab:dataset-stats}
\end{table}
{\small
\begin{table*}
\begin{tabular}{l|c|c|c|c}
$\uparrow$ -- higher is better; $\downarrow$ -- lower is better
&\textbf{Autopilot} 
& \textbf{\begin{tabular}[c]{@{}c@{}}Auto-sklearn \\ ensemble \\ defaults off\end{tabular}} 
& \textbf{\begin{tabular}[c]{@{}c@{}}Auto-sklearn \\  ensemble \\ defaults on\end{tabular}} 
& \textbf{\begin{tabular}[c]{@{}c@{}}XGBoost \\ custom FE \end{tabular}} 
\\
\hline
\textbf{Job Success Rate \hfill $\uparrow$}
&93.2\% 
&74.4 \% 
&82.4\% 
&100.0\%
\\
\textbf{\% matching the baseline \hfill $\uparrow$}
&78.7\%	
&74.0\%	
&91.0\%	
&100.0\%
\\
\textbf{RED, relative error difference \hfill $\downarrow$} 	
&$-10.8 (\pm4.29)\%$ 
&$-7.9  (\pm4.67)\%$
&$-15.2 (\pm4.08)\%$
&$+0.0  (\pm0.0 )\%$
\\
\textbf{Best models produced \hfill $\uparrow$} 	
&$78$	
&$24$	
&$80$	
&$23$
\\
\textbf{Best models produced, no ensemble \hfill $\uparrow$}
&97	
&53
&-
&37
\\
\textbf{Average AutoML Job Runtime\hfill $\downarrow$}
&12119.0 s	
&36011.2 s	
&36123.9 s	
&19.9 s
\\
\textbf{Average Model Endpoint Latency\hfill $\downarrow$}
&97.8 ms	
&76.4 ms	
&939.2 ms	
&38.8 ms
\\
\end{tabular}
\vspace{0.3cm} 
\caption{Performance of Autopilot compared to chosen baseline algorithms.
Reference for the relative error as well as for the percentage of matching error
is the performance of XGBoost.} %
\label{tab:results-comparison}
\end{table*}
}

We compare Autopilot with auto-sklearn \cite{NIPS2015_5872} in two settings and XGBoost \cite{chen2016xgb}.  
Custom feature engineering (FE) is used with XGBoost to handle text or categorical input. 
It first detects the type of each column using simple set of heuristics: 
if all feature values can be converted to numerical values, then the feature is considered to
be of numeric type; if it has less than 20 unique values, it is considered to be
of categorical type, and if there are English words present in the values of the
feature - it is considered to be of text type.
In other cases, the feature is ignored.
For numerical features, missing values are imputed with the mean, and values are
standardized to have variance of one. 
For categorical features, one-hot-encoding is used.
TF-IDF is used for text features.

Our benchmarks include the following algorithms:
\begin{itemize}
\item \textbf{Autopilot} \\ 
Autopilot run with a budget of 250 candidates, and a maximum runtime of 10 hours.

\item \textbf{auto-sklearn, ensemble and defaults on} \\
Auto-sklearn has an option to warm-start Bayesian Optimization using default hyper-parameter configurations which performed well on other datasets, found by a meta-learning step. After training is complete, auto-sklearn can also automatically create an ensemble of models selected with Bayesian Optimization to further improve the performances.

\item \textbf{auto-sklearn, ensemble and defaults off} \\ 
Same as above, except that ensemble construction and metalearning initialization (defaults) of the hyperparameter optimization procedure is disabled, to access the effect of such features on performance and stability.

\item \textbf{XGBoost with custom feature engineering} \\ 
Running XGBoost with single hyperparameter configuration, using equivalent of the latest version of feature preprocessing in Autopilot. One of the purposes of this step is to validate that it is possible for an AutoML procedure to succeed on all of the datasets used for our evaluation. Additionally, performance with such approach indicates if there is any value in running a hyperparameter procedure as used in Autopilot.

\end{itemize}

All of the metrics used for evaluation of the service were designed in such a way that they generalize across regression and classification. To surface different aspects of performance of Autopilot, the following metrics are calculated:
\begin{itemize}

\item \textbf{Relative Error Difference}\\
Difference in normalized error metric. The error metric is the error rate for
classification and RMSE for regression. It is defined as $(A-B)/\max(A, B)$, where $A$ is a score of method, and $B$ is a score of a baseline.
The relative errors are averaged across all datasets.
This metric is affected by small changes in absolute values of the evaluated and baseline methods.

\item \textbf{Job Success Rate} \\
How many of the jobs successfully completed end-to-end, meaning that an ML model was produced, and it was possible to score the model.

\item \textbf{\% matching baseline} \\
How many datasets did the evaluated AutoML approach match or improve upon the baseline method.

\item \textbf{Best models produced} \\
How many times did the AutoML approach produce the best performing model. Note that the numbers need not necessarily sum up to total number of datasets, as some automl approaches might produce equivalent models, as well as for some datasets all the methods compared failed. Note that this metric depends less on small changes in the error metrics.

\item \textbf{Avg. inference endpoint latency} \\
This metric corresponds to the time it takes to process a request to a server, where request contains a batch of 100 rows of test set data, and server performs inferences with the best found model. For the baselines compared to the Autopilot, a Flask application was used to serve the best found models; Autopilot model was deployed as a SageMaker endpoint. The latency of the SageMaker endpoint also includes overhead due to authentication. This metric allows to compare latency change with different model types produced by AutoML.
 
\end{itemize}

A comparison of Autopilot against baselines can be found in Table
\ref{tab:results-comparison}. Note that there is a large variety of datasets in the collection used for the benchmark, and hence it is not trivial to ensure that an AutoML succeeds for all possible edge cases of data. For this reason, we do not achieve perfect success rate with the current version of the Autopilot service, which is a subject of an ongoing work.
Aside from this, we see that Autopilot outperforms the naive baseline of XGBoost with default hyper-parameter values and performs on par with the baselines we have used, even when no ensemble construction is used, which is currently not supported in Autopilot.
The table shows that ensembles provide a boost to predictive performance, but at the expense of training time, model complexity and inference latency. 

%% file: acknowledgments.tex
\section*{Acknowledgments}
Autopilot has contributions from several members  of the SageMaker team,  notably Lakshmi Ramakrishnan, Kumar Venkateswar, Stefano Stefani, Tushar Saxena, Yuqing Gao, Peter Liu, Enrico Sartorello, Furkan Bozkurt, Ugur Adiguzel, Anne Milbert, Choucri Bechir, Per De Silva,  Sylvia Wang, Pei Xu,  Vikram Rajashekharan, Isabel Panapento.

%% file: main.bbl
\begin{thebibliography}{46}
\providecommand{\natexlab}[1]{#1}
\providecommand{\url}[1]{\texttt{#1}}
\expandafter\ifx\csname urlstyle\endcsname\relax
  \providecommand{\doi}[1]{doi: #1}\else
  \providecommand{\doi}{doi: \begingroup \urlstyle{rm}\Url}\fi

\bibitem[ssl(2019)]{ssle}
sagemaker-scikit-learn-extension.
\newblock \url{https://github.com/aws/sagemaker-scikit-learn-extension/}, 2019.

\bibitem[kag(2020)]{kagglewebsite}
Kaggle.com.
\newblock \url{http://kaggle.com}, 2020.

\bibitem[Auer et~al.(2002)Auer, Cesa-Bianchi, Freund, and
  Schapire]{auer2002nonstochastic}
Auer, P., Cesa-Bianchi, N., Freund, Y., and Schapire, R.~E.
\newblock The nonstochastic multiarmed bandit problem.
\newblock \emph{SIAM journal on computing}, 32\penalty0 (1):\penalty0 48--77,
  2002.

\bibitem[Bergstra et~al.(2015)Bergstra, Komer, Eliasmith, Yamins, and
  Cox]{bergstra2015hyperopt}
Bergstra, J., Komer, B., Eliasmith, C., Yamins, D., and Cox, D.
\newblock {Hyperopt}: A {Python} library for model selection and hyperparameter
  optimization.
\newblock \emph{Computational Science and Discovery}, 8\penalty0 (1), 2015.

\bibitem[Buitinck et~al.(2013)Buitinck, Louppe, Blondel, Pedregosa, Mueller,
  Grisel, Niculae, Prettenhofer, Gramfort, Grobler, Layton, VanderPlas, Joly,
  Holt, and Varoquaux]{sklearn_api}
Buitinck, L., Louppe, G., Blondel, M., Pedregosa, F., Mueller, A., Grisel, O.,
  Niculae, V., Prettenhofer, P., Gramfort, A., Grobler, J., Layton, R.,
  VanderPlas, J., Joly, A., Holt, B., and Varoquaux, G.
\newblock {API} design for machine learning software: experiences from the
  scikit-learn project.
\newblock In \emph{ECML PKDD Workshop: Languages for Data Mining and Machine
  Learning}, pp.\  108--122, 2013.

\bibitem[Caruana et~al.(2004)Caruana, Niculescu-Mizil, Crew, and
  Ksikes]{caruana2004icml}
Caruana, R., Niculescu-Mizil, A., Crew, G., and Ksikes, A.
\newblock Ensemble selection from libraries of models.
\newblock In \emph{Proceedings of the 21st International Conference on Machine
  Learning}, 2004.

\bibitem[Chen \& Guestrin(2016)Chen and Guestrin]{chen2016xgb}
Chen, T. and Guestrin, C.
\newblock Xgboost: A scalable tree boosting system.
\newblock In \emph{Proceedings of the 22nd ACM SIGKDD international conference
  on Knowledge discovery and data mining}, pp.\  785--794, 2016.

\bibitem[DataRobot()]{datarobot_automl}
DataRobot.
\newblock \emph{Automated Machine Learning}.
\newblock \url{https://www.datarobot.com/platform/automated-machine-learning/}.

\bibitem[Dua \& Graff(2017)Dua and Graff]{Dua:2019}
Dua, D. and Graff, C.
\newblock {UCI} machine learning repository, 2017.
\newblock URL \url{http://archive.ics.uci.edu/ml}.

\bibitem[Erickson et~al.(2019)Erickson, Mueller, Shirkov, Zhang, and
  Li]{auto_gluon}
Erickson, N., Mueller, J., Shirkov, A., Zhang, H., and Li, M.
\newblock \emph{AutoGluon}, 2019.
\newblock \url{https://autogluon.mxnet.io/}.

\bibitem[Erickson et~al.(2020)Erickson, Mueller, Shirkov, Zhang, Larroy, Li,
  and Smola]{agtabular}
Erickson, N., Mueller, J., Shirkov, A., Zhang, H., Larroy, P., Li, M., and
  Smola, A.
\newblock Autogluon-tabular: Robust and accurate automl for structured data.
\newblock \emph{arXiv preprint arXiv:2003.06505}, 2020.

\bibitem[Falkner et~al.(2018)Falkner, Klein, and Hutter]{falkner2018bohb}
Falkner, S., Klein, A., and Hutter, F.
\newblock Bohb: Robust and efficient hyperparameter optimization at scale.
\newblock \emph{arXiv preprint arXiv:1807.01774}, 2018.

\bibitem[Feurer et~al.(2015)Feurer, Klein, Eggensperger, Springenberg, Blum,
  and Hutter]{NIPS2015_5872}
Feurer, M., Klein, A., Eggensperger, K., Springenberg, J., Blum, M., and
  Hutter, F.
\newblock Efficient and robust automated machine learning.
\newblock In Cortes, C., Lawrence, N.~D., Lee, D.~D., Sugiyama, M., and
  Garnett, R. (eds.), \emph{Advances in Neural Information Processing Systems
  28}, pp.\  2962--2970. 2015.

\bibitem[Feurer et~al.(2019)Feurer, Klein, Eggensperger, Springenberg, Blum,
  and Hutter]{feurer2019auto}
Feurer, M., Klein, A., Eggensperger, K., Springenberg, J.~T., Blum, M., and
  Hutter, F.
\newblock {Auto-sklearn}: efficient and robust automated machine learning.
\newblock In \emph{Automated Machine Learning}, pp.\  113--134. Springer, 2019.

\bibitem[Gijsbers \& Vanschoren(2019)Gijsbers and Vanschoren]{gijsbers2019gama}
Gijsbers, P. and Vanschoren, J.
\newblock {GAMA}: Genetic automated machine learning assistant.
\newblock \emph{Journal of Open Source Software}, 4\penalty0 (33), 2019.

\bibitem[Google(2019)]{google_automl}
Google.
\newblock \emph{AutoML Tables}, 2019.
\newblock \url{https://cloud.google.com/automl-tables/}.

\bibitem[Guinet et~al.(2020)Guinet, Perrone, and Archambeau]{guinet2020}
Guinet, G., Perrone, V., and Archambeau, C.
\newblock {Pareto-efficient Acquisition Functions for Cost-Aware Bayesian
  Optimization}.
\newblock \emph{NeurIPS Meta Learning Workshop}, 2020.

\bibitem[H2O.ai(2017)]{h2o_automl}
H2O.ai.
\newblock \emph{H2O AutoML}, 2017.
\newblock \url{ http://docs.h2o.ai/h2o/latest-stable/h2o-docs/automl.html}.

\bibitem[He et~al.(2019)He, Zhao, and Chu]{he2019automl}
He, X., Zhao, K., and Chu, X.
\newblock Automl: A survey of the state-of-the-art.
\newblock \emph{arXiv preprint arXiv:1908.00709}, 2019.

\bibitem[Hutter et~al.(2009)Hutter, Hoos, Leyton-Brown, and
  St{\"u}tzle]{hutter2009paramils}
Hutter, F., Hoos, H.~H., Leyton-Brown, K., and St{\"u}tzle, T.
\newblock Paramils: an automatic algorithm configuration framework.
\newblock \emph{Journal of Artificial Intelligence Research}, 36:\penalty0
  267--306, 2009.

\bibitem[Hutter et~al.(2010)Hutter, Hoos, and
  Leyton-Brown]{hutter2010sequential}
Hutter, F., Hoos, H.~H., and Leyton-Brown, K.
\newblock Sequential model-based optimization for general algorithm
  configuration (extended version).
\newblock \emph{Technical Report TR-2010--10, University of British Columbia,
  Computer Science, Tech. Rep.}, 2010.

\bibitem[Jin et~al.(2019)Jin, Song, and Hu]{jin2019autokeras}
Jin, H., Song, Q., and Hu, X.
\newblock {Auto-keras}: An efficient neural architecture search system.
\newblock In \emph{Proceedings of the 25th ACM SIGKDD International Conference
  on Knowledge Discovery and Data Mining}, pp.\  1946--1956, 2019.

\bibitem[Lee et~al.(2020)Lee, Perrone, Archambeau, and Seeger]{Lee2020}
Lee, E.~H., Perrone, V., Archambeau, C., and Seeger, M.
\newblock Cost-aware {Bayesian} optimization.
\newblock In \emph{ICML AutoML Workshop}, 2020.

\bibitem[Microsoft(2019)]{microsoft_automl}
Microsoft.
\newblock \emph{Azure AutoML}, 2019.
\newblock
  \url{https://azure.microsoft.com/en-us/services/machine-learning/automatedml/}.

\bibitem[Olson \& Moore(2019)Olson and Moore]{olson2019tpot}
Olson, R.~S. and Moore, J.~H.
\newblock Tpot: A tree-based pipeline optimization tool for automating machine
  learning.
\newblock In \emph{Automated Machine Learning}, pp.\  151--160. Springer, 2019.

\bibitem[Pandey(2019)]{pandey2019h2o}
Pandey, P.
\newblock A deep dive into {H2Os AutoML}.
\newblock Technical report, 2019.
\newblock URL \url{http://www.h2o.ai/blog/a-deep- dive-into-h2os-automl/}.

\bibitem[Pedregosa et~al.(2011)Pedregosa, Varoquaux, Gramfort, Michel, Thirion,
  Grisel, Blondel, Prettenhofer, Weiss, Dubourg, Vanderplas, Passos,
  Cournapeau, Brucher, Perrot, and Duchesnay]{scikit-learn}
Pedregosa, F., Varoquaux, G., Gramfort, A., Michel, V., Thirion, B., Grisel,
  O., Blondel, M., Prettenhofer, P., Weiss, R., Dubourg, V., Vanderplas, J.,
  Passos, A., Cournapeau, D., Brucher, M., Perrot, M., and Duchesnay, E.
\newblock Scikit-learn: Machine learning in {P}ython.
\newblock \emph{Journal of Machine Learning Research}, 12:\penalty0 2825--2830,
  2011.

\bibitem[Perrone et~al.(2017)Perrone, Jenatton, Seeger, and
  Archambeau]{perrone_multiple_2017}
Perrone, V., Jenatton, R., Seeger, M., and Archambeau, C.
\newblock Multiple adaptive {Bayesian} linear regression for scalable
  {Bayesian} optimization with warm start.
\newblock \emph{NeurIPS Meta Learning Workshop}, 2017.

\bibitem[Perrone et~al.(2018)Perrone, Jenatton, Seeger, and
  Archambeau]{Perrone2018}
Perrone, V., Jenatton, R., Seeger, M.~W., and Archambeau, C.
\newblock Scalable hyperparameter transfer learning.
\newblock \emph{Advances in Neural Information Processing Systems 31}, pp.\
  6845--6855, 2018.

\bibitem[Perrone et~al.(2019{\natexlab{a}})Perrone, Shcherbatyi, Jenatton,
  Archambeau, and Seeger]{Perrone2019mes}
Perrone, V., Shcherbatyi, I., Jenatton, R., Archambeau, C., and Seeger, M.
\newblock Constrained {Bayesian} optimization with max-value entropy search.
\newblock In \emph{NeurIPS Meta Learning Workshop}, 2019{\natexlab{a}}.

\bibitem[Perrone et~al.(2019{\natexlab{b}})Perrone, Shen, Seeger, Archambeau,
  and Jenatton]{Perrone2019}
Perrone, V., Shen, H., Seeger, M.~W., Archambeau, C., and Jenatton, R.
\newblock {Learning search spaces for Bayesian optimization: Another view of
  hyperparameter transfer learning}.
\newblock \emph{Advances in Neural Information Processing Systems},
  32:\penalty0 12771--12781, 2019{\natexlab{b}}.

\bibitem[Perrone et~al.(2020)Perrone, Donini, Kenthapadi, and
  Archambeau]{Perrone2020}
Perrone, V., Donini, M., Kenthapadi, K., and Archambeau, C.
\newblock Fair {Bayesian} optimization.
\newblock In \emph{ICML AutoML Workshop}, 2020.

\bibitem[Salesforce(2018)]{transmogrifai_automl}
Salesforce.
\newblock \emph{TransmogrifAI}, 2018.
\newblock \url{https://transmogrif.ai/}.

\bibitem[Salinas et~al.(2020)Salinas, Shen, and Perrone]{Salinas2020}
Salinas, D., Shen, H., and Perrone, V.
\newblock A quantile-based approach for hyperparameter transfer learning.
\newblock \emph{International Conference on Machine Learning}, pp.\
  7706--7716, 2020.

\bibitem[Seznec et~al.(2020)Seznec, Menard, Lazaric, and
  Valko]{seznec2020single}
Seznec, J., Menard, P., Lazaric, A., and Valko, M.
\newblock A single algorithm for both restless and rested rotting bandits.
\newblock In \emph{International Conference on Artificial Intelligence and
  Statistics}, pp.\  3784--3794, 2020.

\bibitem[Sparkcognition(2019)]{darwin_automl}
Sparkcognition.
\newblock \emph{Darwin AutoML}, 2019.
\newblock \url{https://www.sparkcognition.com/products/darwin/}.

\bibitem[Sutton et~al.(1998)Sutton, Barto, et~al.]{sutton1998introduction}
Sutton, R.~S., Barto, A.~G., et~al.
\newblock \emph{Introduction to reinforcement learning}, volume 135.
\newblock MIT press Cambridge, 1998.

\bibitem[Thornton et~al.(2013)Thornton, Hutter, Hoos, and
  Leyton-Brown]{thornton2013auto}
Thornton, C., Hutter, F., Hoos, H.~H., and Leyton-Brown, K.
\newblock {Auto-WEKA}: Combined selection and hyperparameter optimization of
  classification algorithms.
\newblock In \emph{Proceedings of the 19th ACM SIGKDD international conference
  on Knowledge discovery and data mining}, pp.\  847--855, 2013.

\bibitem[van Rijn et~al.(2018)van Rijn, Pfisterer, Thomas, Muller, Bischl, and
  Vanschoren]{van2018meta}
van Rijn, J.~N., Pfisterer, F., Thomas, J., Muller, A., Bischl, B., and
  Vanschoren, J.
\newblock Meta learning for defaults--symbolic defaults.
\newblock In \emph{Neural Information Processing Workshop on Meta-Learning},
  2018.

\bibitem[Vanschoren(2018)]{vanschoren2018meta}
Vanschoren, J.
\newblock Meta-learning: A survey.
\newblock \emph{arXiv preprint arXiv:1810.03548}, 2018.

\bibitem[Vanschoren et~al.(2013)Vanschoren, van Rijn, Bischl, and
  Torgo]{OpenML2013}
Vanschoren, J., van Rijn, J.~N., Bischl, B., and Torgo, L.
\newblock Openml: Networked science in machine learning.
\newblock \emph{SIGKDD Explorations}, 15\penalty0 (2):\penalty0 49--60, 2013.
\newblock \doi{10.1145/2641190.2641198}.
\newblock URL \url{http://doi.acm.org/10.1145/2641190.2641198}.

\bibitem[Wang et~al.(2019)Wang, Weisz, Muller, Ram, Geyer, Dugan, Tausczik,
  Samulowitz, and Gray]{Wang2019}
Wang, D., Weisz, J.~D., Muller, M., Ram, P., Geyer, W., Dugan, C., Tausczik,
  Y., Samulowitz, H., and Gray, A.
\newblock Human-ai collaboration in data science: Exploring data scientists'
  perceptions of automated ai.
\newblock \emph{Proceedings of the ACM on Human-Computer Interaction}, 2019.

\bibitem[Winkelmolen et~al.(2020)Winkelmolen, Ivkin, Bozkurt, and
  Karnin]{winkelmolen2020practical}
Winkelmolen, F., Ivkin, N., Bozkurt, H.~F., and Karnin, Z.
\newblock Practical and sample efficient zero-shot hpo.
\newblock \emph{arXiv preprint arXiv:2007.13382}, 2020.

\bibitem[Wistuba et~al.(2015{\natexlab{a}})Wistuba, Schilling, and
  Schmidt-Thieme]{wistuba2015learning}
Wistuba, M., Schilling, N., and Schmidt-Thieme, L.
\newblock Learning hyperparameter optimization initializations.
\newblock In \emph{2015 IEEE international conference on data science and
  advanced analytics (DSAA)}, pp.\  1--10. IEEE, 2015{\natexlab{a}}.

\bibitem[Wistuba et~al.(2015{\natexlab{b}})Wistuba, Schilling, and
  Schmidt-Thieme]{wistuba2015sequential}
Wistuba, M., Schilling, N., and Schmidt-Thieme, L.
\newblock Sequential model-free hyperparameter tuning.
\newblock In \emph{2015 IEEE international conference on data mining}, pp.\
  1033--1038. IEEE, 2015{\natexlab{b}}.

\bibitem[Z{\"o}ller \& Huber(2019)Z{\"o}ller and Huber]{zoller2019survey}
Z{\"o}ller, M.-A. and Huber, M.~F.
\newblock Survey on automated machine learning.
\newblock \emph{arXiv preprint arXiv:1904.12054}, 9, 2019.

\end{thebibliography}
